\documentclass[letterpaper]{article} 
\usepackage{aaai2027} 
\usepackage[hyphens]{url} 
\usepackage{graphicx} 
\usepackage{natbib} 
\usepackage{caption} 
\usepackage{algorithm}
\usepackage{algorithmic}
\usepackage{amsmath}
\usepackage{booktabs}
\usepackage{cuted}
\usepackage{xcolor}
\definecolor{targetred}{HTML}{C00000}
\definecolor{targetblue}{HTML}{4471C4}

\newcommand{\methodname}{\textsc{AffectDelta}}
\newcommand{\datasetname}{\textsc{AffectPair-249K}}

\title{AffectDelta: Beyond Emotion Labels for Image Editing}
\author{Xingzu Zhan, Lin Gu\textsuperscript{\rm 2}, Ruogu Fang\textsuperscript{\rm 1}}
\affiliations{\textsuperscript{\rm 1}Department of Biomedical Engineering, Vanderbilt University, Nashville, Tennessee, USA\\ \textsuperscript{\rm 2}Research Institute of Electrical Communication, Tohoku University, Sendai, Japan\\ xingzuz@andrew.cmu.edu, lin.gu@riken.jp, ruogu.fang@vanderbilt.edu}

\begin{document}
\maketitle

\begin{strip}
    \centering
    \includegraphics[width=0.90\textwidth]{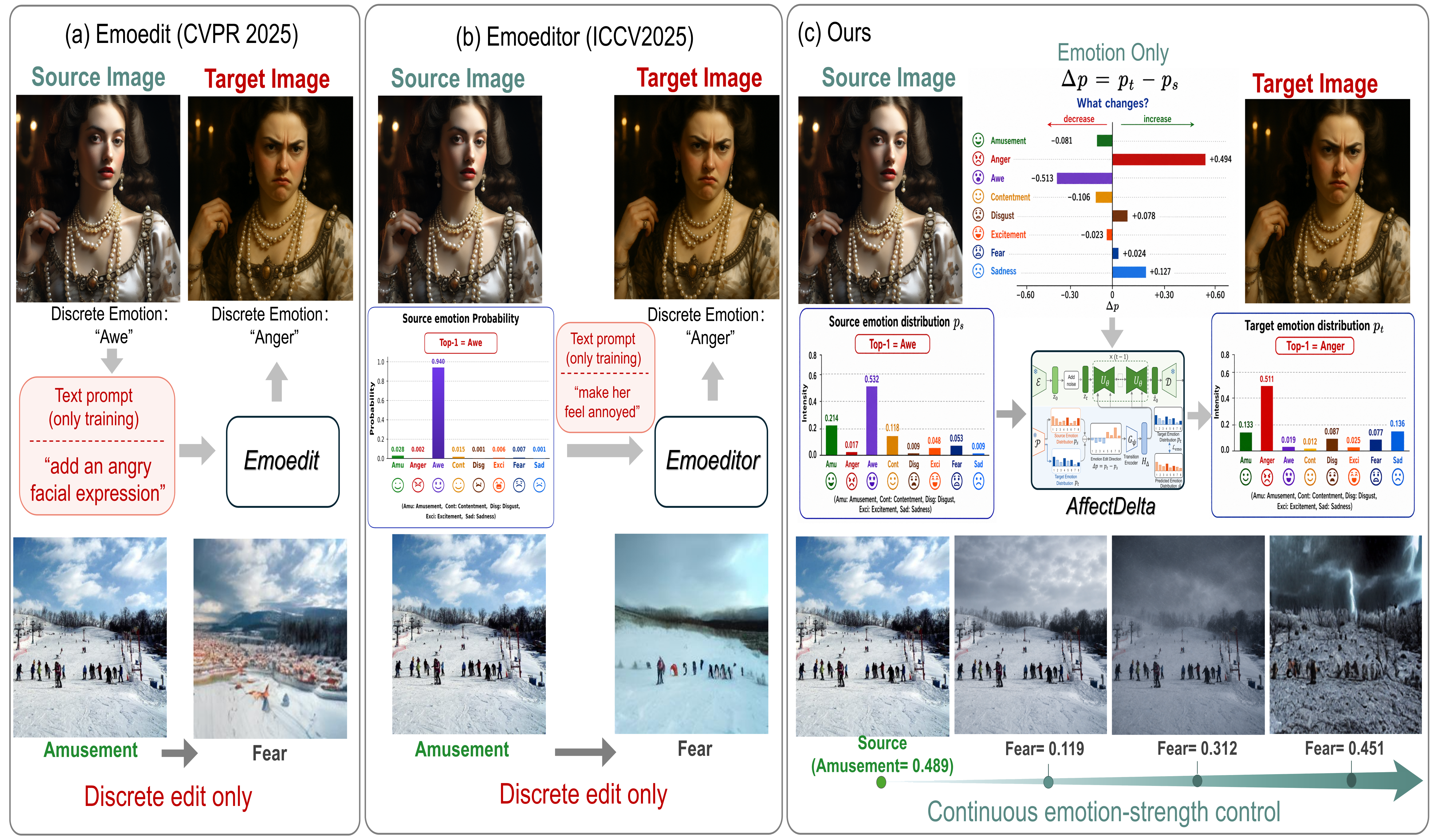}
    \vspace{-2mm}
\captionof{figure}{Motivation and conceptual comparison. (a) EmoEdit conditions its Emotion Adapter on the source image and a single target-emotion label, but does not explicitly represent source affect; during training, its emotion embedding is aligned with a content-editing instruction. (b) EmoEditor applies an eight-way emotion classifier to the source and subtracts its class posterior from a one-hot target code. Trained for single-label classification rather than emotion-distribution learning, this predictor tends to yield sharply concentrated outputs; the derived editing direction is likewise aligned with instruction text during training. (c) \methodname{} instead represents both endpoints as full eight-dimensional emotion distributions and conditions the editor on their signed difference, while paired images supervise context-dependent visual realizations without editing-text alignment.}
    \label{fig:overview}
    \vspace{-2mm}
\end{strip}

\begin{abstract}
Emotion-driven image editing aims to evoke a specified target emotion by modifying emotion-relevant visual cues in a source image, while preserving the overall composition and semantic--structural coherence of the original scene. Existing scene-level editors typically specify the target with a single emotion category and often learn visual transformations from operation-level text instructions. A category collapses a mixed affective endpoint into one dominant label, while language cannot precisely quantify how coexisting emotions should increase, decrease, or remain stable. We introduce \methodname{}, a source-aware editor that treats editing as a transition between eight-dimensional emotion distributions. A frozen Emotion Distribution Predictor estimates the source state, and the signed source-to-target difference encodes the direction and magnitude of the requested transition. Within \methodname{}, an internal transition encoder and a source-aware diffusion backbone jointly translate this signal into context-dependent semantic and appearance changes. To train this formulation, we construct \datasetname{}, comprising 248,841 source--target pairs with predicted eight-dimensional distributions and spanning both cross-category and within-category transitions. Experiments against six baselines, combining quantitative evaluation with qualitative comparisons, demonstrate improved affective alignment and content preservation, while ablations validate our design choices.
Code and dataset will be made publicly available upon acceptance.\end{abstract}
\section{Introduction}

Psychological research has long used images as controlled affective stimuli, demonstrating that visual input produces measurable changes in valence and arousal \cite{bradley2007international}. Yet these responses are not determined by image content alone: appraisal theory shows that emotion depends on how viewers interpret a situation \cite{smith1985patterns}, and mixed-emotion studies further show that positive and negative feelings can co-occur \cite{larsen2001people}. Consistent with this view, a single image may evoke several emotions in the same viewer rather than one dominant category \cite{peng2015mixed}, while large-scale visual-affect studies attribute such responses to the interaction of semantic content, visual attributes, and subjective interpretation \cite{achlioptas2021artemis,yang2023emoset}.\par These findings motivate \emph{emotion-driven image editing}: modifying emotion-relevant cues to steer the affect evoked by an image while preserving the scene identity and content unrelated to the intended change. Such control could support creative design, advertising, personalized visual communication, and the construction of controlled affective stimuli for psychological research \cite{behnke2026using,chiu2026beyond}. Achieving it, however, requires an editor to represent not only the desired affective state but also the mixture of emotions already elicited by the source image.

Recent scene-level emotion editors have advanced from global color and style adjustment to semantic modifications of objects, actions, and scenes~\cite{lin2025make,yang2025emoedit,zhang2026emokgedit}. These advances show that effective editing must select context-dependent visual cues, but the representative pipelines in Fig.~\ref{fig:overview}(a--b) expose two unresolved challenges. First, a single target label cannot represent the full affective transition from a source image that already evokes a mixture of emotions. EmoSet notes that one image may evoke several emotions even though its annotations assign one dominant category~\cite{yang2023emoset}. EmoEditor estimates the source distribution but defines the destination as a one-hot category, while EmoEdit and EmoKGEdit are likewise driven by a target category or emotion word~\cite{lin2025make,yang2025emoedit,zhang2026emokgedit}. A category therefore identifies only the dominant endpoint, leaving the source mixture and the component-wise direction and magnitude of change unspecified, particularly when source and target retain the same Top-1 emotion. Second, the visual realization of an affective transition is often mediated by language during training or inference. Language-driven editors such as Affective Image Filter, InstructPix2Pix, and Language-Driven Artistic Style Transfer require textual control at inference, whereas EmoEditor and EmoEdit use editing or content instructions as language-alignment targets during training~\cite{weng2023affective,brooks2023instructpix2pix,fu2022language,lin2025make,yang2025emoedit}. Such instructions can name a mood or prescribe one plausible operation, but they do not explicitly encode how each emotion component should change. Moreover, the same affective transition may require different visual realizations across scenes~\cite{sun2023imagebrush}.

The two challenges point to the same missing interface: an affective control signal that quantifies the source-to-target change without prescribing a visual operation. We therefore introduce \methodname{}, a unified editor that formulates emotional image editing as a transition between full distributions over eight emotion categories, as summarized in Fig.~\ref{fig:overview}(c). Following the categorical model of Mikels et al.~\cite{mikels2005emotional}, the signed difference between the source and target distributions explicitly represents which emotions should increase or decrease and by how much. A frozen \emph{Emotion Distribution Predictor} (EDP) grounds images in this distribution space and, together with task-specific filtering, enables us to construct \datasetname{}, containing 248,841 source--target pairs that span both cross-category transitions and same-Top-1 distribution shifts. During adaptation, these paired visual outcomes supervise context-dependent realizations of each transition without editing-text conditioning or affect--text alignment. Within \methodname{}, an internal transition encoder maps the signed difference to conditioning tokens, which a source-aware diffusion backbone combines with the source-image representation to generate semantic and appearance changes. The distribution difference thus specifies \emph{what affective change is requested}, while the paired images teach the unified editor \emph{how to realize it in the current scene}. Against six external baselines, \methodname{} ranks first on all six metrics; controlled ablations further support full-distribution targets and signed source-to-target conditioning, and show that the tested text-alignment pathway is not required in our setting.

Our contributions are threefold:
\begin{itemize}
    \item We formulate scene-level emotional image editing as a signed transition between full distributions over eight emotion categories, preserving mixed source and target states while explicitly representing the direction and magnitude of the requested change beyond a single label.
    \item We construct \datasetname{}, a distribution-aware dataset of 248,841 task-filtered source--target pairs that covers both cross-category transitions and same-Top-1 distribution shifts, and establish a paired visual learning setting without editing-text supervision during adaptation.
    \item We develop \methodname{}, a unified source-aware diffusion editor that jointly learns transition encoding and scene-dependent visual realization. Experiments against six external baselines show the best performance across all six metrics, while controlled ablations support the full-distribution target, signed source-to-target difference, and direct affective conditioning.
\end{itemize}

\section{Related Work}

\subsection{Visual Emotion Representation}
Visual emotion analysis has evolved from single-label categorization to distributional modeling of affective ambiguity and subjectivity. Psychological norms defined categorical emotions~\cite{mikels2005emotional}, while early methods linked them to low-level and semantic cues~\cite{machajdik2010affective,borth2013large}; web benchmarks then scaled category recognition~\cite{you2016building}. Yet a dominant label discards observer disagreement and secondary emotions. Emotion6 introduced population-level distributions~\cite{peng2015mixed}, and Flickr-LDL and Twitter-LDL extended them to the eight-dimensional Mikels space~\cite{yang2017learning}, inspiring joint category--distribution and relation-aware models~\cite{yang2017joint,yang2021circular}. Later datasets added affective explanations or interpretable attributes~\cite{achlioptas2021artemis,yang2023emoset}. We instead condition semantic editing directly on the signed difference between source and target distributions.

\subsection{Affective Image Editing}
Early affective editing primarily changed appearance. Distribution-driven methods transferred color and texture from exemplars or retrieved references, including control with a user-specified seven-dimensional distribution~\cite{peng2015mixed,ali2017automatic}. Affective Image Filter and AIF-D instead map emotional text to appearance transformations~\cite{weng2023affective,zhang2026deeper}. These approaches enabled distribution- or language-guided control but largely preserved emotion-relevant scene semantics.

Scene-level generative editors instead modify objects and actions. EmoEditor is closest to our formulation: it subtracts a predicted source distribution from a target code, but uses a one-hot destination and learns edit embeddings from target-emotion-driven instructions despite text-free inference~\cite{lin2025make}. EmoEdit and AIEdiT likewise derive editing semantics from constructed instructions or emotional descriptions~\cite{yang2025emoedit,zhang2026aiedit}. EmoAgent, Moodifier, and EmoKGEdit translate target emotions into plans, prompts, or localized cues~\cite{mao2026emoagent,ye2025moodifier,zhang2026emokgedit}, whereas MooD retrieves affective references for continuous valence--arousal control~\cite{yin2026mood}. Thus, fine-grained editing semantics still commonly come from language or references. Our editor instead adapts from paired images and their full eight-dimensional source and target distributions, without textual editing instructions. Their signed difference explicitly encodes which emotions should increase or decrease and by how much, while the visual pairs supervise the scene-dependent realization.

\subsection{Instruction-Guided Image Editing}
Diffusion editors preserve source structure through noisy-guide denoising or cross-attention control~\cite{meng2022sdedit,hertz2023prompt}. Instruction-following models map an image and free-form command directly to an edit, with MGIE enriching terse commands using a multimodal language model~\cite{brooks2023instructpix2pix,sheynin2024emu,fu2024mgie}. Training data evolved from MagicBrush's manually annotated triplets to UltraEdit's millions of real-image-anchored, region-annotated examples~\cite{zhang2023magicbrush,zhao2024ultraedit}. ImageBrush instead specifies operations through visual exemplar pairs~\cite{sun2023imagebrush}. Nevertheless, commands and exemplars specify content- or style-level operations, not calibrated multidimensional affective transitions. We instead condition editing on a signed emotion-distribution change, leaving its visual realization to the source scene.

\section{Method}
\begin{figure*}[t]
    \centering
    \includegraphics[width=0.90\textwidth]{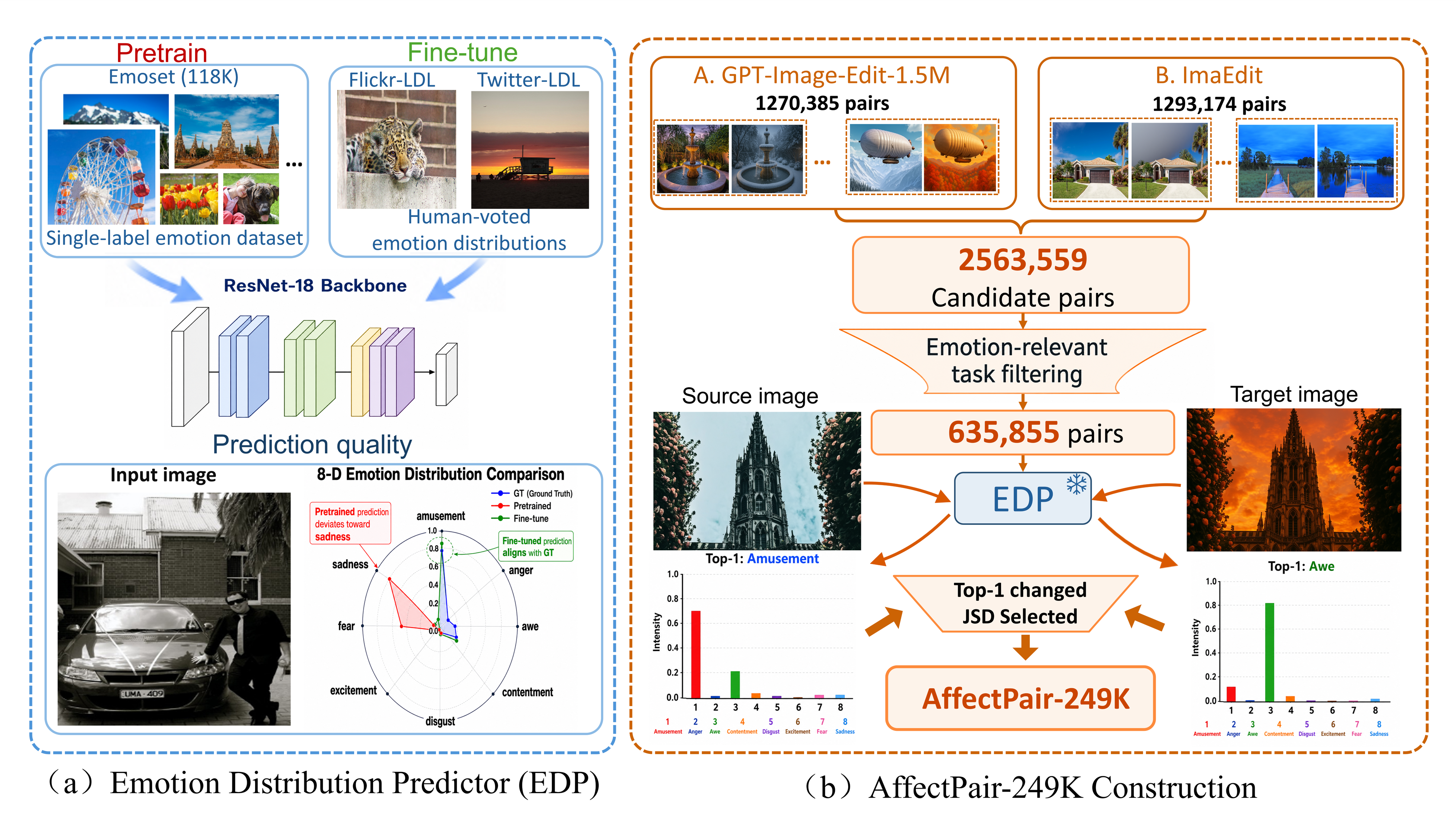}
    \vspace{-5mm}
    \caption{Emotion prediction and pair construction. (a) The EDP is pretrained on EmoSet with single-label supervision and fine-tuned on Flickr-LDL and Twitter-LDL with human-vote distributions. (b) Candidate edit pairs first undergo task-specific emotion-relevance filtering; EDP-based Top-1 and JSD selection then retains cross-category and within-category affective transitions for \datasetname.}
    \label{fig:dataset-construction}
    \vspace{-5mm}
\end{figure*}

\subsection{Overview and Problem Formulation}

We formulate emotional image editing as an affective transition rather than category-conditioned generation. Let $I_s$ be a source image and let $\mathbf{p}_t$ be the desired distribution over $K=8$ emotions: amusement, anger, awe, contentment, disgust, excitement, fear, and sadness, following the categorical space of Mikels et al.~\cite{mikels2005emotional}. Each distribution lies on the probability simplex, so its entries are nonnegative and sum to one. A frozen Emotion Distribution Predictor (EDP) $P_\phi$ estimates the affect already evoked by the source, $\mathbf{p}_s=P_\phi(I_s)$. We denote \methodname{} by $F_\Theta$ and represent the requested edit as
\begin{equation}
\Delta\mathbf{p}=\mathbf{p}_t-\mathbf{p}_s,
\qquad
\widehat I=F_\Theta(I_s,\Delta\mathbf{p}).
\label{eq:transition}
\end{equation}
Positive and negative entries specify which emotions should increase or decrease, while their magnitudes quantify the requested change. Because both endpoints are normalized, $\sum_k\Delta p_k=0$. This representation retains the dominant and secondary emotions of both endpoints instead of reducing either endpoint to an isolated category.

The frozen EDP and \methodname{} play distinct roles. The EDP grounds images in the eight-dimensional affective space and annotates paired images for dataset construction. \methodname{} is the learnable editor: its internal transition encoder $G_\psi$ maps $\Delta\mathbf{p}$ into transition tokens, and its source-aware diffusion backbone $U_\theta$ combines them with the source-image latent to learn a scene-dependent visual realization, where $\Theta=\{\psi,\theta\}$. During training, the paired target supplies $\mathbf{p}_t$, and an auxiliary emotion loss regularizes a denoised image estimate toward this endpoint. At inference, $\mathbf{p}_t$ is the desired endpoint supplied to the editor. No operation-level text instruction is used as an explicit conditioning signal.

\begin{figure*}[t]
    \centering
    \includegraphics[width=0.85\textwidth]{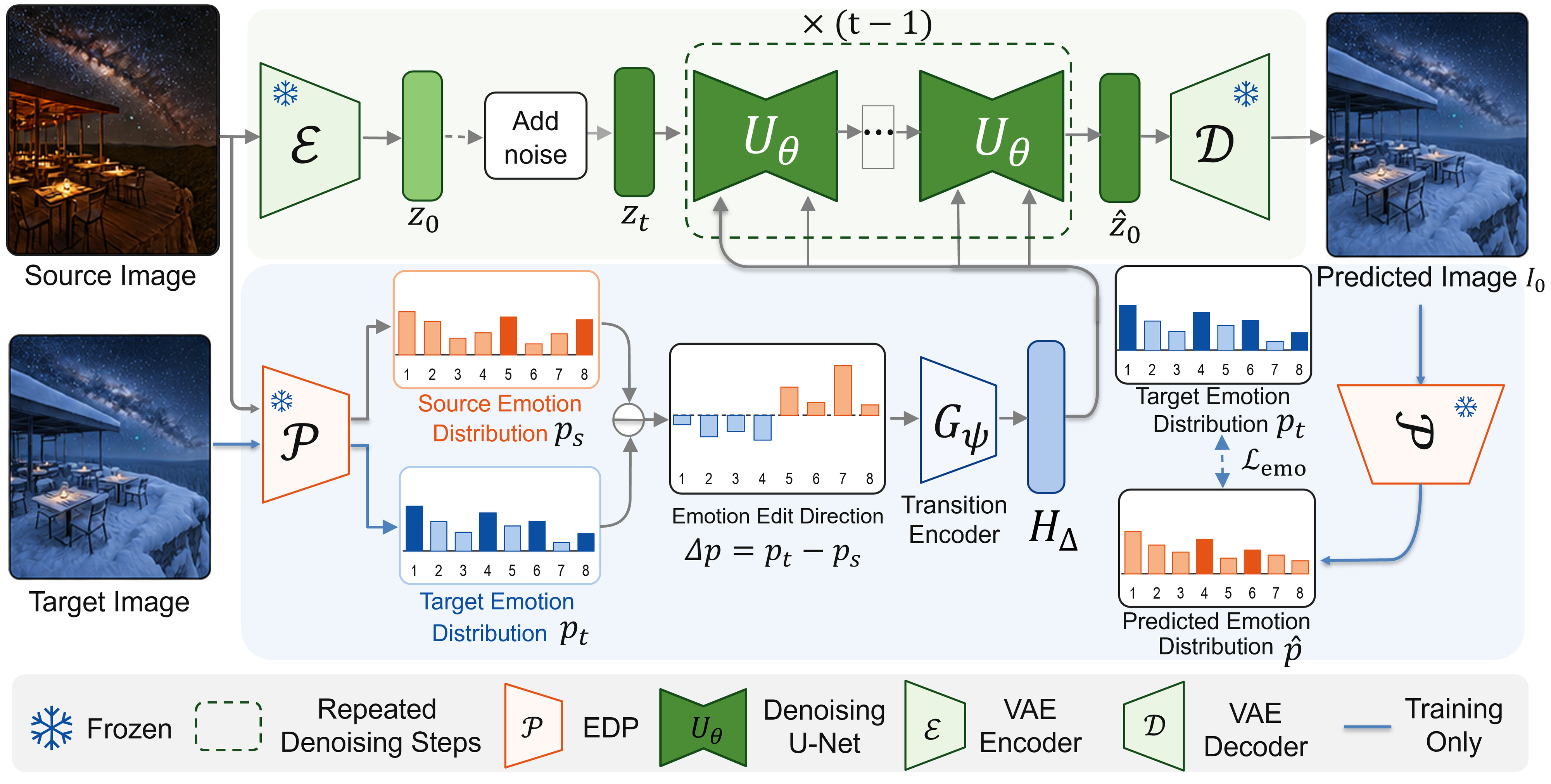}
    \vspace{-2mm}
    \caption{\textbf{Architecture of \methodname{}.}
    The frozen EDP $P_\phi$ (shown as $P$) estimates the source and target distributions $\mathbf{p}_s$ and $\mathbf{p}_t$; their signed difference $\Delta\mathbf{p}$ is encoded by $G_\psi$ into transition tokens $\mathbf{H}_{\Delta}$ that condition the denoising U-Net $U_\theta$. At inference, the source latent is noised, iteratively denoised under this condition, and decoded into the edited image. During training, the frozen EDP evaluates the reconstructed prediction, and $\mathcal{L}_{\mathrm{emo}}$ aligns its distribution $\widehat{\mathbf{p}}$ with $\mathbf{p}_t$.}
    \label{fig:affectdelta-architecture}
    \vspace{-5mm}
\end{figure*}

\subsection{Emotion Distribution Predictor}

The EDP converts an image into a complete affective distribution. We instantiate $P_\phi$ with a ResNet-18~\cite{he2016deep} whose classification head produces eight logits followed by a softmax. As illustrated in Fig.~\ref{fig:dataset-construction}(a), training proceeds in two stages. We first pretrain the network on EmoSet~\cite{yang2023emoset} using its single-label annotations over the same eight emotion categories. We then fine-tune the final residual stage and prediction head on Flickr-LDL and Twitter-LDL, whose human-vote annotations form distributions in this emotion space~\cite{yang2017learning}. For an image $I_i$ with label distribution $\mathbf{q}_i$, the distribution-level fine-tuning objective is
\begin{equation}
\mathcal{L}_{\mathrm{EDP}}
=\frac{1}{B}\sum_{i=1}^{B}
D_{\mathrm{KL}}\!\left(\mathbf{q}_i\,\Vert\,P_\phi(I_i)\right).
\label{eq:edp}
\end{equation}
This second stage replaces hard-label supervision with distributional supervision, preserving the relative mass assigned by annotators to coexisting emotions. Consequently, this fine-tuning turns $P_\phi$ from a single-label classifier of the emotion expressed by an image into an image-evoked emotion distribution predictor that estimates the relative intensities of the eight emotions elicited in human observers.

\subsection{Distribution-Aware Pair Construction}

As summarized in Fig.~\ref{fig:dataset-construction}(b), we construct \datasetname through task-specific emotion-relevance filtering followed by distribution-aware selection. We begin with 2,563,559 candidate source--target pairs: 1,270,385 from GPT-Image-Edit-1.5M~\cite{wang2025gpt} and 1,293,174 from ImgEdit~\cite{ye2025imgedit}. The task-specific filter removes pairs that are unsuitable for learning affective transitions, including emotion-irrelevant object substitutions and other edits that do not meaningfully change the scene's affective content. This stage retains 635,855 emotion-relevant pairs.

For each remaining pair $(I_s,I_t)$, the frozen EDP predicts $\mathbf{p}_s=P_\phi(I_s)$ and $\mathbf{p}_t=P_\phi(I_t)$, from which we compute $\Delta\mathbf{p}$ using Eq.~(\ref{eq:transition}). We then retain two complementary transition types. Cross-category pairs have different predicted Top-1 emotions, capturing changes in the dominant affect. Within-category pairs preserve the Top-1 emotion but exhibit a large Jensen--Shannon divergence,
\begin{equation}
\begin{aligned}
\mathbf{m}
&= \tfrac{1}{2}\bigl(\mathbf{p}_s+\mathbf{p}_t\bigr),\\
\mathrm{JSD}(\mathbf{p}_s,\mathbf{p}_t)
&= \tfrac{1}{2}D_{\mathrm{KL}}(\mathbf{p}_s\Vert\mathbf{m})
 + \tfrac{1}{2}D_{\mathrm{KL}}(\mathbf{p}_t\Vert\mathbf{m}).
\end{aligned}
\label{eq:jsd}
\end{equation}
The within-category criterion retains affective changes that categorical supervision would discard, such as strengthening the dominant emotion or rebalancing secondary emotions without changing the dominant category. Together, these criteria produce \datasetname with 248,841 pairs spanning cross-category transitions and within-category distribution shifts.

Each retained example is represented as $(I_s,I_t,\mathbf{p}_s,\mathbf{p}_t,\Delta\mathbf{p})$. Although the source corpora provide editing instructions, we exclude them from editor optimization. Training uses only the paired images, $\Delta\mathbf{p}$, and $\mathbf{p}_t$, so the visual difference between source and target directly teaches the model how changes in objects, attributes, and scene context reshape the evoked emotion distribution. Rather than binding an affective transition to a particular linguistic description or predefined operation, the editor learns from visual outcomes how that transition should be realized in each source scene.

\subsection{\methodname{} Architecture}

As illustrated in Fig.~\ref{fig:affectdelta-architecture}, \methodname{} is the complete editing model $F_\Theta$. It jointly represents the requested affective change and realizes it in the source scene through an internal transition encoder $G_\psi$ and a source-aware diffusion backbone $U_\theta$, which are optimized end to end.

The transition encoder $G_\psi$ uses a four-layer MLP to project the eight-dimensional transition vector $\Delta\mathbf{p}$ into a $77\times768$ cross-attention representation. Its output comprises 77 transition tokens of dimension 768:
\begin{equation}
\mathbf{H}_{\Delta}=G_\psi(\Delta\mathbf{p}),
\qquad
\mathrm{shape}(\mathbf{H}_{\Delta})=77\times768.
\label{eq:transition_encoding}
\end{equation}
This shape matches the pretrained backbone's text-conditioning interface, allowing the transition tokens to replace text embeddings.

The source-aware diffusion backbone retains the latent image-conditioning pathway of InstructPix2Pix~\cite{brooks2023instructpix2pix}, which is built on latent diffusion~\cite{rombach2022high}, and replaces language embeddings with $\mathbf{H}_{\Delta}$. Let $\mathcal{E}$ and $\mathcal{D}$ denote the frozen VAE encoder and decoder, and let $s$ be the VAE scaling factor. For a training pair, the target image provides the clean denoising latent $\mathbf{z}_0=s\,\mathcal{E}(2I_t-1)$, while the source image provides an unscaled conditioning latent $\mathbf{c}_s=\mathcal{E}(2I_s-1)$. At a randomly sampled diffusion step $t$, Gaussian noise $\epsilon$ is added as
\begin{equation}
\mathbf{z}_t=\sqrt{\bar\alpha_t}\,\mathbf{z}_0
+\sqrt{1-\bar\alpha_t}\,\epsilon.
\label{eq:forward_diffusion}
\end{equation}
The denoising U-Net receives the channel-wise concatenation of the noisy target latent and source condition, while its cross-attention layers receive the transition tokens:
\begin{equation}
\widehat{\epsilon}
=U_\theta\!\left([\mathbf{z}_t;\mathbf{c}_s],t,\mathbf{H}_{\Delta}\right).
\label{eq:editor_denoising}
\end{equation}
The VAE remains frozen, whereas the two trainable parts of \methodname{}, $G_\psi$ and $U_\theta$, are adapted jointly. The source latent anchors scene structure, while the transition tokens specify where affect should move.

Notably, our editor adaptation introduces no text supervision: training neither conditions on editing instructions nor aligns affect representations with text embeddings. Instead, the signed emotion-distribution transition directly conditions the editor, while the paired source--target images teach \methodname{} how that transition should reshape scene content and appearance. \methodname{} therefore learns visual realizations of affective transitions rather than binding emotions to specific editing instructions.

At inference, the source latent is noised and iteratively denoised under the same visual condition and $\mathbf{H}_{\Delta}$, so generation requires only $I_s$ and the desired affective endpoint.

\subsection{Training Objectives}

We train \methodname{} with the standard diffusion noise-prediction objective. Given a paired sample $(I_s,I_t)$, a diffusion timestep $t$, and Gaussian noise $\epsilon$, the objective is
\begin{equation}
\mathcal{L}_{\mathrm{diff}}
=\mathrm{E}_{I_s,I_t,t,\epsilon}
\left[\left\|\epsilon-\widehat{\epsilon}\right\|_2^2\right].
\label{eq:diffusion_loss}
\end{equation}
This objective teaches the editor to reproduce the visual transformation represented by each source--target pair, but it does not explicitly require the edited image to match the target affect distribution. We therefore add an auxiliary emotion loss, $\mathcal{L}_{\mathrm{emo}}$, evaluated by the frozen EDP. On an emotion-rebalanced subset $\mathcal{B}_e$ and a low-noise timestep $t_e$, the U-Net predicts $\widehat{\epsilon}_{t_e}$, from which we reconstruct a differentiable one-step estimate of the clean target latent,
\begin{equation}
\widehat{\mathbf{z}}_0
=
\frac{\mathbf{z}_{t_e}-\sqrt{1-\bar\alpha_{t_e}}\,
\widehat{\epsilon}_{t_e}}
{\sqrt{\bar\alpha_{t_e}}}.
\label{eq:pred_x0}
\end{equation}
For numerical stability, we clip $\widehat{\mathbf{z}}_0$ to $[-c,c]$, with $c=10$ in our implementation. We then divide the clipped latent by the VAE scaling factor $s$, decode it with the frozen VAE $\mathcal{D}$, and rescale the decoder output to $[0,1]$ to obtain the image estimate $\widehat I_0$. The frozen EDP predicts the emotion distribution of each reconstructed sample as $\widehat{\mathbf{p}}_i=P_\phi(\widehat I_{0,i})$. This auxiliary reconstruction is used only to construct $\mathcal{L}_{\mathrm{emo}}$ during training and is not the inference output of \methodname{}. We align each predicted distribution with its corresponding target distribution using
\begin{equation}
\begin{array}{rcl}
\mathcal{L}_{\mathrm{emo}}
&=&\frac{1}{|\mathcal{B}_e|K}
\displaystyle\sum_{i\in\mathcal{B}_e}\\
&&\left\|\widehat{\mathbf{p}}_i-\mathbf{p}_{t,i}\right\|_2^2,\\[1pt]
\mathcal{L}&=&\mathcal{L}_{\mathrm{diff}}
+\lambda_{\mathrm{emo}}\mathcal{L}_{\mathrm{emo}}.
\end{array}
\label{eq:total_loss}
\end{equation}
The one-step estimate avoids running a complete sampling trajectory inside every training iteration. The diffusion loss learns a plausible visual realization from the paired target, whereas $\mathcal{L}_{\mathrm{emo}}$ encourages its affective distribution to match the desired endpoint.

\section{Experiments}

\begin{figure*}[t]
\centering
\begingroup
\setlength{\tabcolsep}{0.7pt}
\renewcommand{\arraystretch}{0.93}

\newcommand{\qimg}[1]{%
  \parbox[c][0.090\textheight][c]{0.118\textwidth}{%
    \centering
    #1%
  }%
}
\newcommand{\qcell}[2]{%
  \shortstack[c]{\qimg{#1}\\[-1.5pt]{\scriptsize #2}}%
}
\newcommand{\qtargetbox}[2]{%
  \parbox[c][0.090\textheight][c]{1.35em}{%
    \centering
    \rotatebox[origin=c]{90}{%
      \textcolor{#1}{\scriptsize\rmfamily\bfseries\itshape ``#2''}%
    }%
  }%
}
\newcommand{\qtarget}[2]{%
  \shortstack[c]{\qtargetbox{#1}{#2}\\[-1.5pt]{\scriptsize\phantom{0.0000}}}%
}
\newcommand{\qplain}[1]{%
  \shortstack[c]{\qimg{#1}\\[-1.5pt]{\scriptsize\phantom{0.0000}}}%
}

\resizebox{0.97\textwidth}{!}{\begin{tabular}{@{}c@{\hspace{0.5pt}}*{8}{c}@{}}
{} &
{\scriptsize\bfseries Input} &
{\scriptsize\bfseries\shortstack{\methodname{}\\[-1pt](Ours)}} &
{\scriptsize\bfseries EmoEditor} &
{\scriptsize\bfseries EmoEdit} &
{\scriptsize\bfseries AIF} &
{\scriptsize\bfseries IP2P} &
{\scriptsize\bfseries SDEdit} &
{\scriptsize\bfseries LDAST}
\\[1pt]

\qtarget{targetred}{Amusement} &
\qplain{\includegraphics[width=\linewidth,height=0.090\textheight,keepaspectratio]{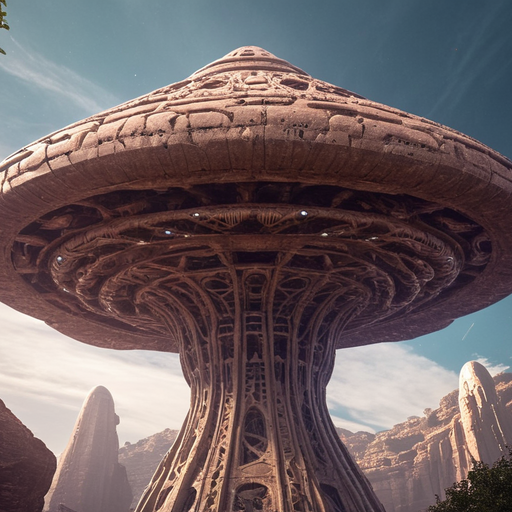}} &
\qcell{\includegraphics[width=\linewidth,height=0.090\textheight,keepaspectratio]{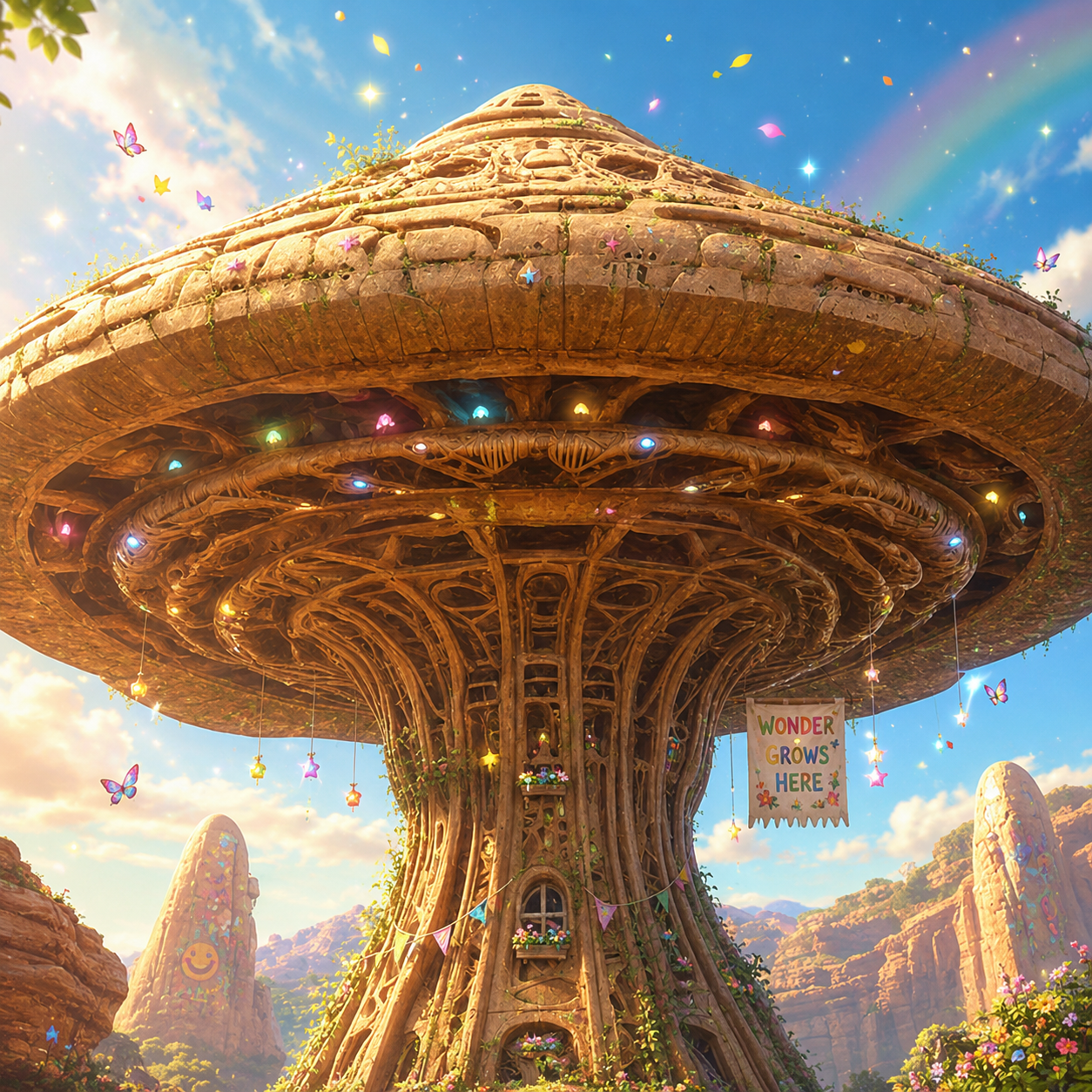}}{\textbf{0.0505}} &
\qcell{\includegraphics[width=\linewidth,height=0.090\textheight,keepaspectratio]{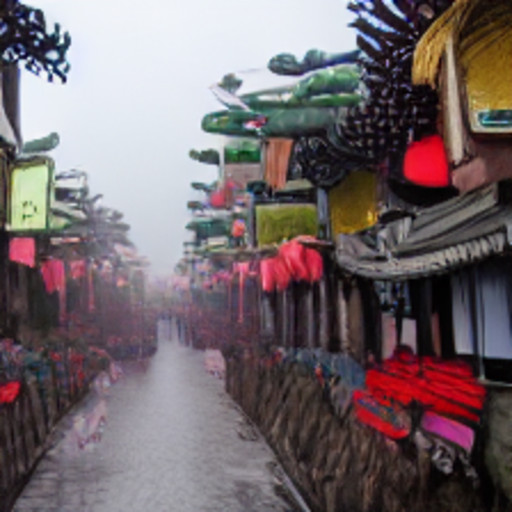}}{0.2479} &
\qcell{\includegraphics[width=\linewidth,height=0.090\textheight,keepaspectratio]{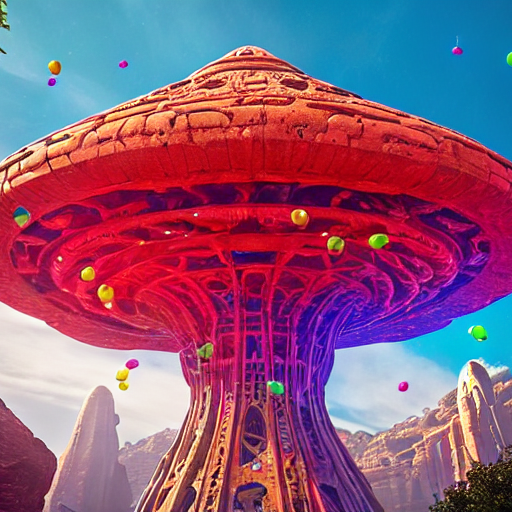}}{0.1155} &
\qcell{\includegraphics[width=\linewidth,height=0.090\textheight,keepaspectratio]{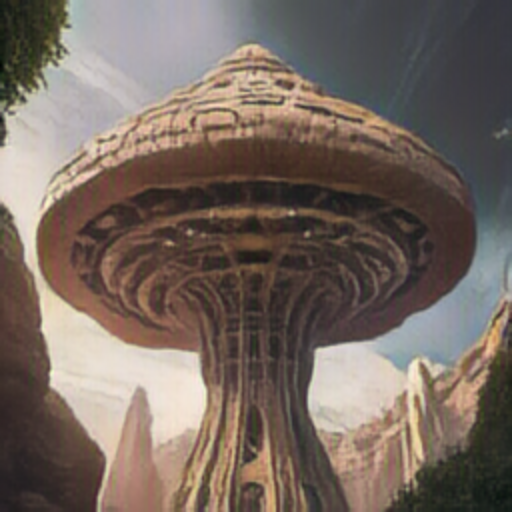}}{0.0577} &
\qcell{\includegraphics[width=\linewidth,height=0.090\textheight,keepaspectratio]{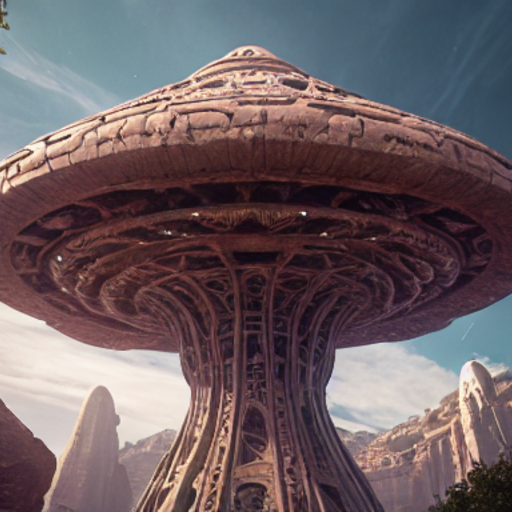}}{0.0976} &
\qcell{\includegraphics[width=\linewidth,height=0.090\textheight,keepaspectratio]{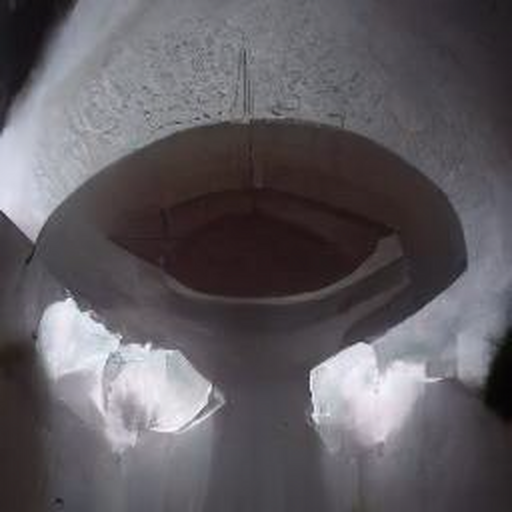}}{0.1606} &
\qcell{\includegraphics[width=\linewidth,height=0.090\textheight,keepaspectratio]{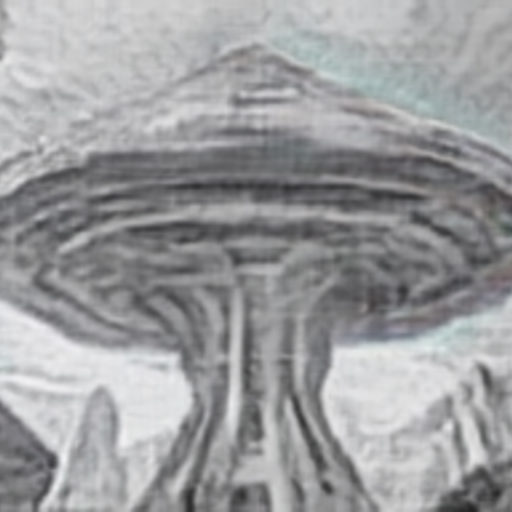}}{0.1138}
\\[-1pt]

\qtarget{targetblue}{Fear} &
\qplain{\includegraphics[width=\linewidth,height=0.090\textheight,keepaspectratio]{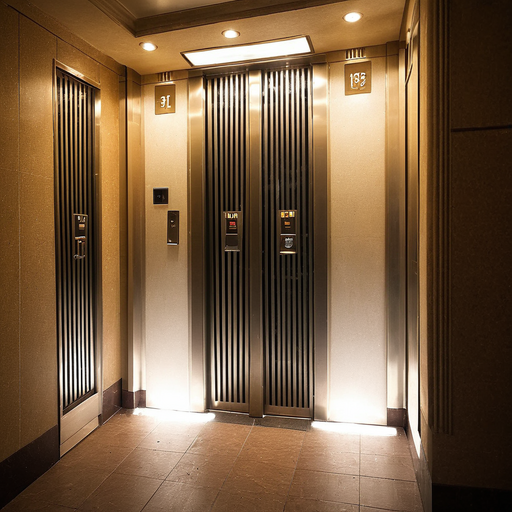}} &
\qcell{\includegraphics[width=\linewidth,height=0.090\textheight,keepaspectratio]{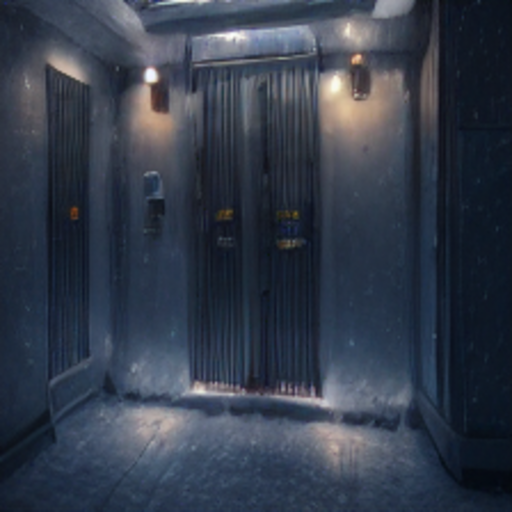}}{\textbf{0.0181}} &
\qcell{\includegraphics[width=\linewidth,height=0.090\textheight,keepaspectratio]{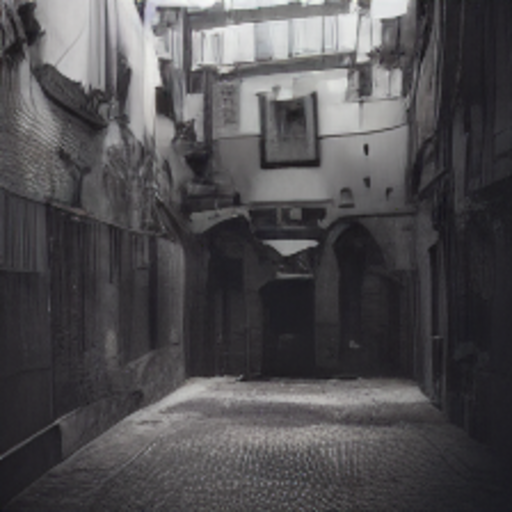}}{0.2113} &
\qcell{\includegraphics[width=\linewidth,height=0.090\textheight,keepaspectratio]{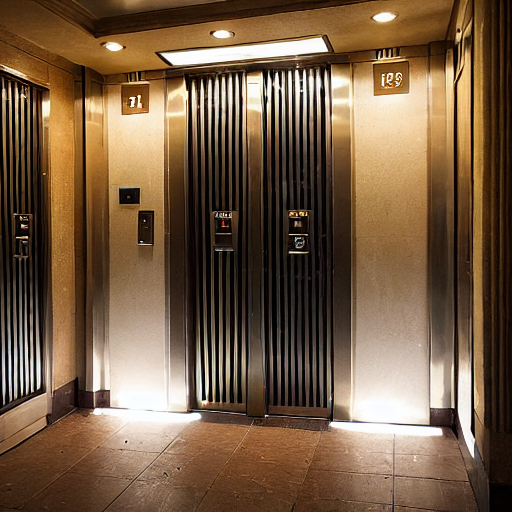}}{0.1119} &
\qcell{\includegraphics[width=\linewidth,height=0.090\textheight,keepaspectratio]{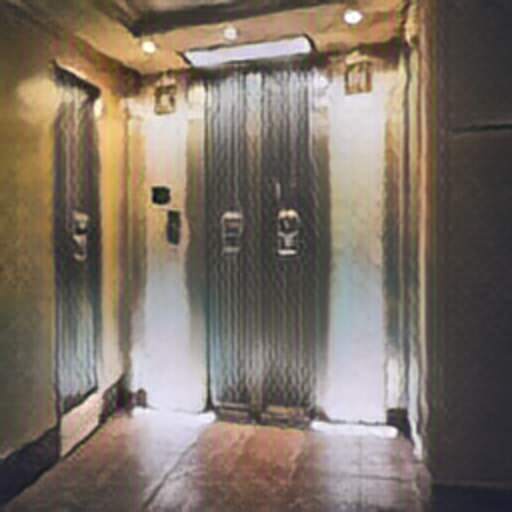}}{0.1273} &
\qcell{\includegraphics[width=\linewidth,height=0.090\textheight,keepaspectratio]{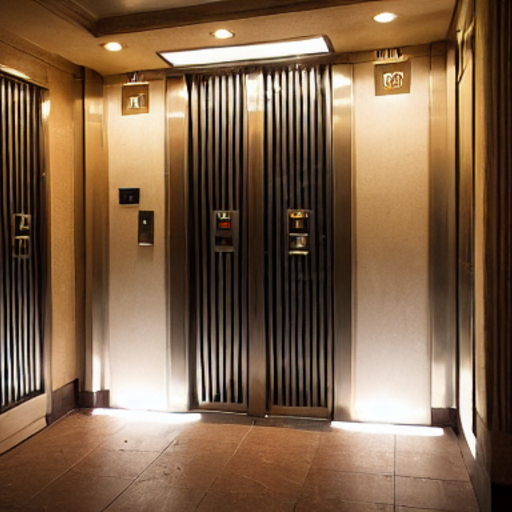}}{0.1814} &
\qcell{\includegraphics[width=\linewidth,height=0.090\textheight,keepaspectratio]{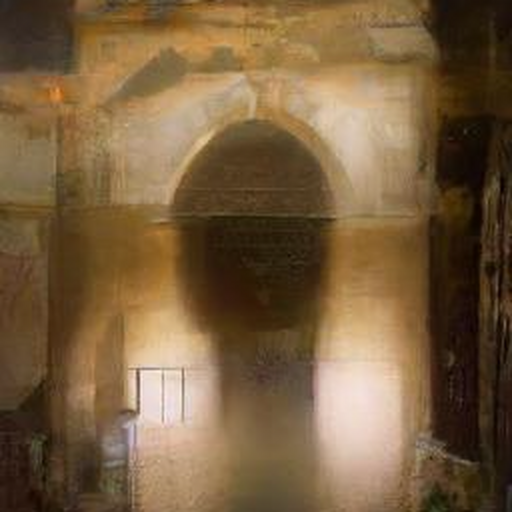}}{0.3855} &
\qcell{\includegraphics[width=\linewidth,height=0.090\textheight,keepaspectratio]{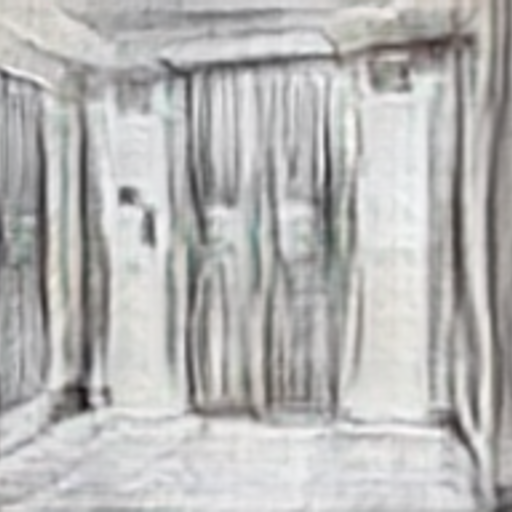}}{0.1663}
\\[-1pt]

\qtarget{targetred}{Awe} &
\qplain{\includegraphics[width=\linewidth,height=0.090\textheight,keepaspectratio]{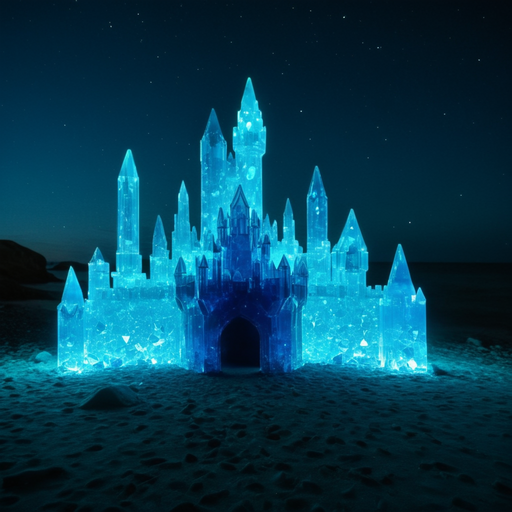}} &
\qcell{\includegraphics[width=\linewidth,height=0.090\textheight,keepaspectratio]{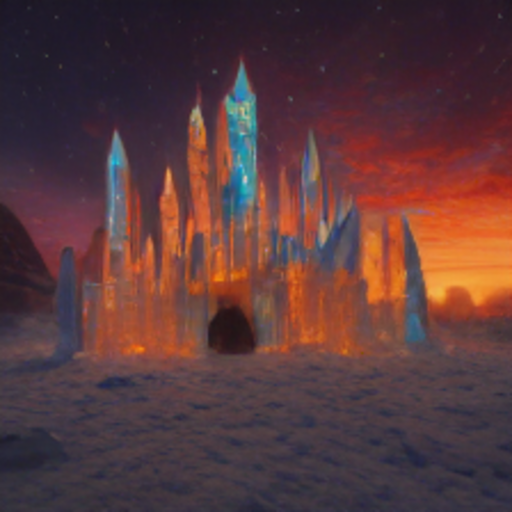}}{\textbf{0.0594}} &
\qcell{\includegraphics[width=\linewidth,height=0.090\textheight,keepaspectratio]{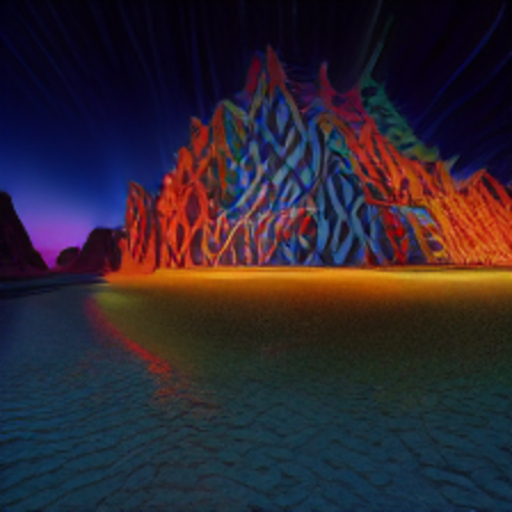}}{0.0599} &
\qcell{\includegraphics[width=\linewidth,height=0.090\textheight,keepaspectratio]{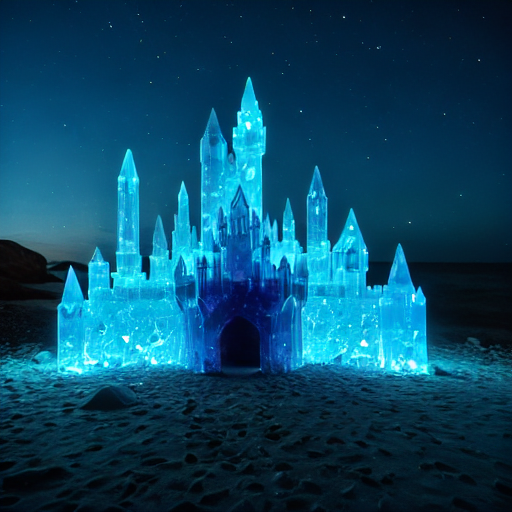}}{0.0790} &
\qcell{\includegraphics[width=\linewidth,height=0.090\textheight,keepaspectratio]{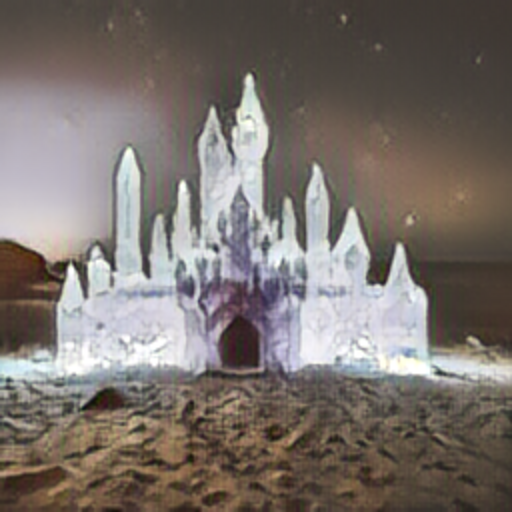}}{0.0791} &
\qcell{\includegraphics[width=\linewidth,height=0.090\textheight,keepaspectratio]{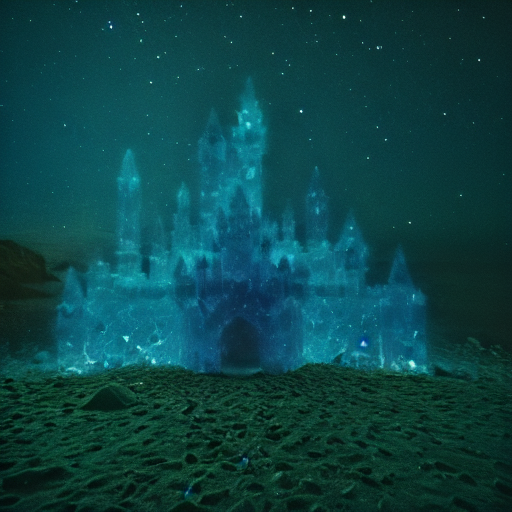}}{0.0711} &
\qcell{\includegraphics[width=\linewidth,height=0.090\textheight,keepaspectratio]{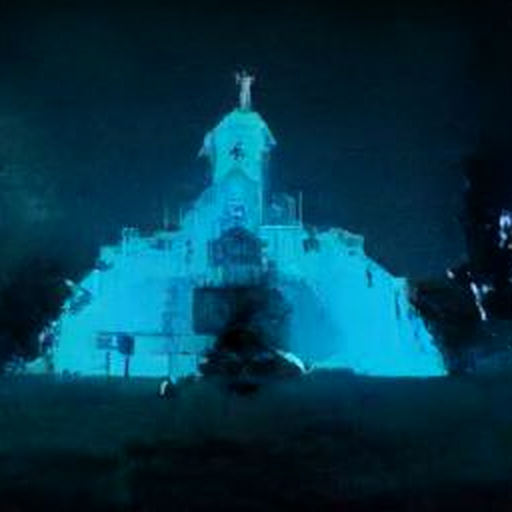}}{0.2062} &
\qcell{\includegraphics[width=\linewidth,height=0.090\textheight,keepaspectratio]{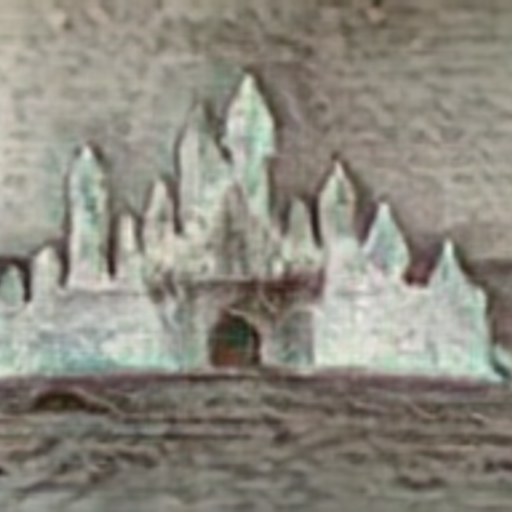}}{0.2460}
\\[-1pt]

\qtarget{targetblue}{Disgust} &
\qplain{\includegraphics[width=\linewidth,height=0.090\textheight,keepaspectratio]{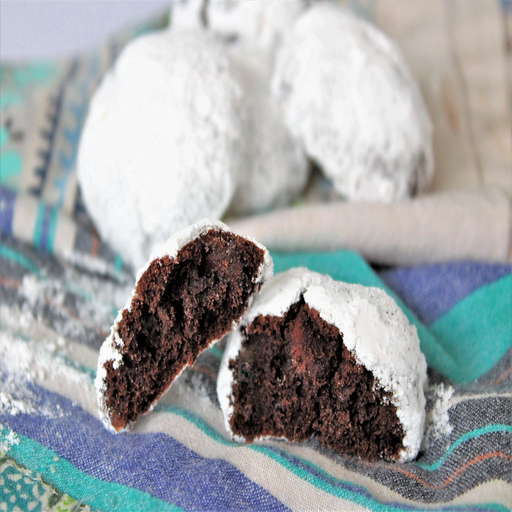}} &
\qcell{\includegraphics[width=\linewidth,height=0.090\textheight,keepaspectratio]{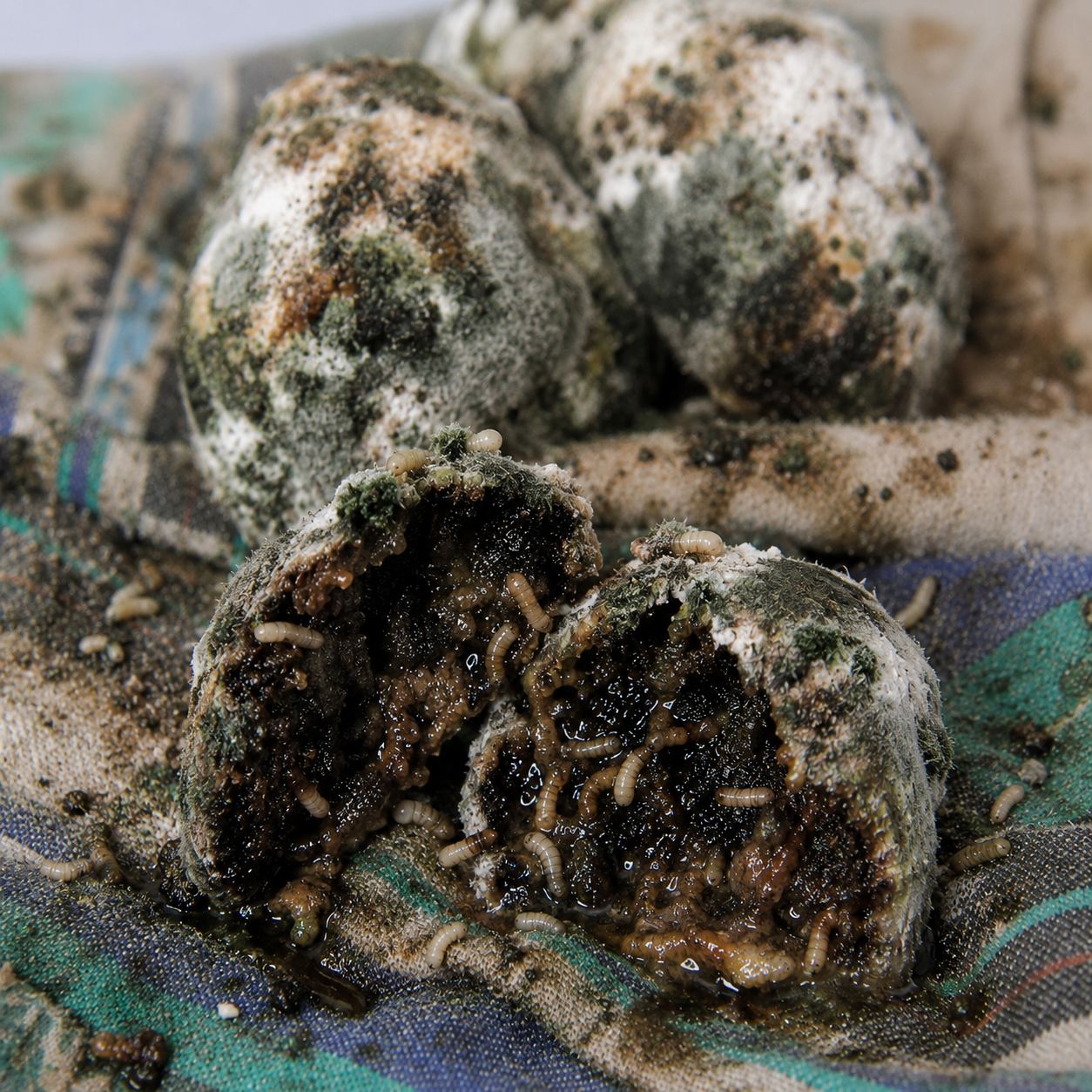}}{\textbf{0.0088}} &
\qcell{\includegraphics[width=\linewidth,height=0.090\textheight,keepaspectratio]{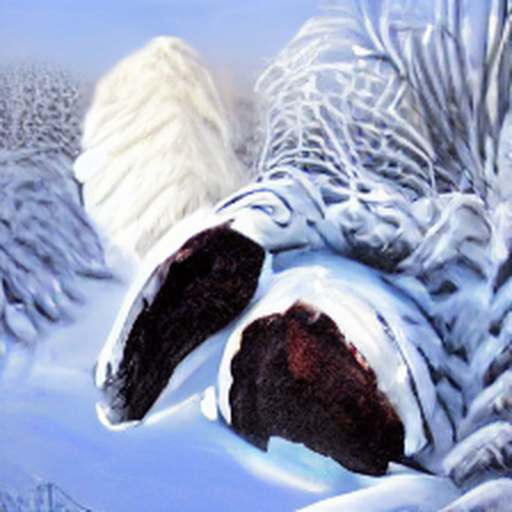}}{0.1293} &
\qcell{\includegraphics[width=\linewidth,height=0.090\textheight,keepaspectratio]{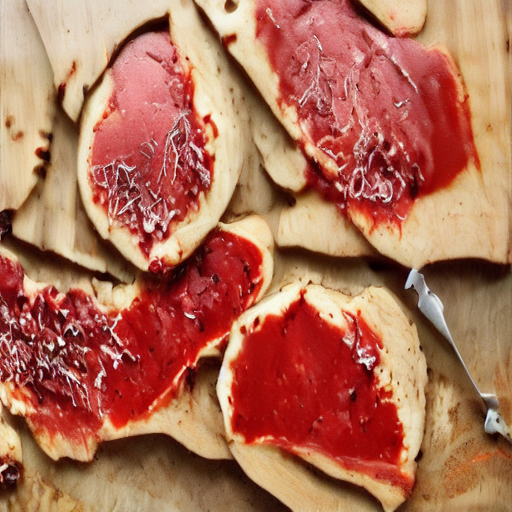}}{0.0552} &
\qcell{\includegraphics[width=\linewidth,height=0.090\textheight,keepaspectratio]{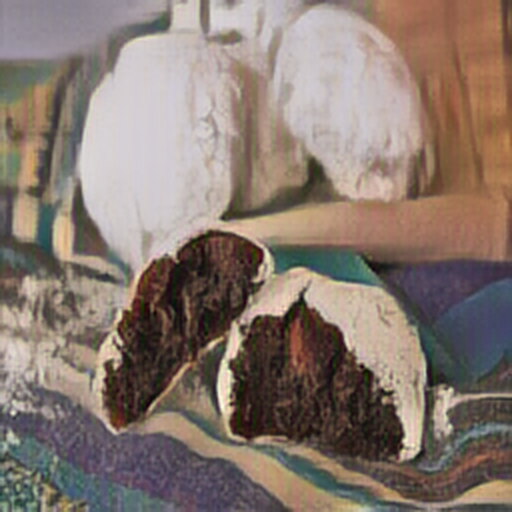}}{0.0893} &
\qcell{\includegraphics[width=\linewidth,height=0.090\textheight,keepaspectratio]{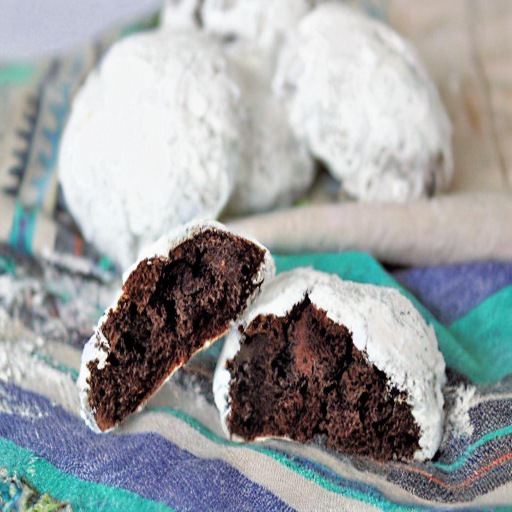}}{0.0128} &
\qcell{\includegraphics[width=\linewidth,height=0.090\textheight,keepaspectratio]{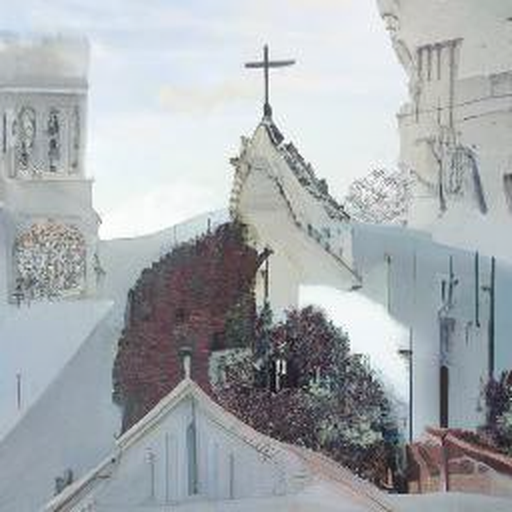}}{0.4447} &
\qcell{\includegraphics[width=\linewidth,height=0.090\textheight,keepaspectratio]{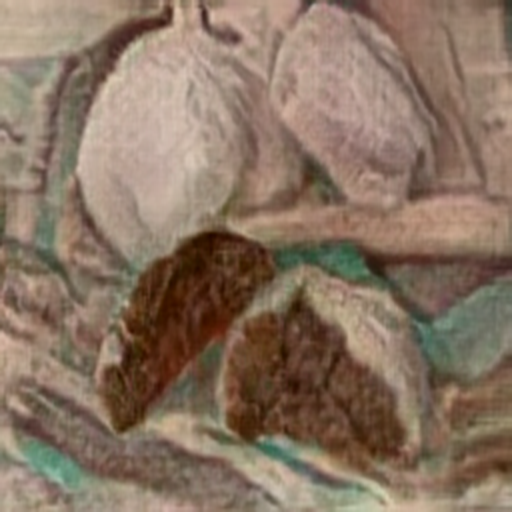}}{0.0324}
\end{tabular}
}\endgroup

\caption{\textbf{Qualitative comparison on four test-set inputs.} Each row shows a source input, its target emotion at left, and outputs from \methodname{} and six baselines. The reported values are the base-2 JSD~$\downarrow$ between the distribution predicted by the independent evaluator and the target distribution; lower is better, and the best score in each row is bold. Across four target emotions, \methodname{} achieves the lowest JSD in every displayed case while realizing scene-dependent semantic and appearance changes.}
\label{fig:qualitative-comparison}
\end{figure*}

\subsection{Experimental Setup}

\paragraph{Data split.}
\datasetname{} contains 248,841 source--target image pairs. We select 4,977 pairs as a disjoint test set, approximately balancing the eight target-emotion categories and retaining both cross-category and same-Top-1 transitions. The remaining 243,864 pairs are used for training. All quantitative and qualitative comparisons are conducted on this test set.

\paragraph{Baselines.}
We compare \methodname{} with six representative editors spanning generic, language-driven, and emotion-specific editing. Affective Image Filter (AIF) reflects emotional text through image appearance~\cite{weng2023affective}; InstructPix2Pix (IP2P) follows operation-level language instructions for semantic editing~\cite{brooks2023instructpix2pix}; and Language-Driven Artistic Style Transfer (LDAST) transfers language-specified artistic attributes~\cite{fu2022language}. These three methods require language at inference. EmoEditor conditions a source-aware diffusion editor on a target emotion category~\cite{lin2025make}, while EmoEdit combines the source image with a target emotion word through an Emotion Adapter~\cite{yang2025emoedit}; neither requires an operation-level instruction at inference, but both use editing or content instructions as language-alignment targets during training. We further include SDEdit as an image-guided diffusion baseline~\cite{meng2022sdedit}. AIEdiT, EmoAgent, Moodifier, EmoKGEdit, and MooD~\cite{zhang2026aiedit,mao2026emoagent,ye2025moodifier,zhang2026emokgedit,yin2026mood} are not compared because no runnable official implementations were publicly available at submission.

\paragraph{Evaluation metrics.}
All affective metrics are computed using an independent emotion-distribution evaluator that is distinct from the EDP used for dataset construction and editor training. Target Top-1 measures whether an output reaches the target's dominant emotion, whereas JSD measures agreement with the complete target distribution. Normalized Affective Progress (NAP) quantifies how far editing advances from the source distribution toward the target, and $\Delta$-Cos measures whether the realized eight-dimensional change follows the requested direction. CLIP-I and LPIPS assess semantic and perceptual preservation with respect to the source image. Together, these metrics distinguish categorical success from full-distribution alignment, source-aware transition fidelity, and content preservation.

\begin{figure*}[t]
    \centering
    \includegraphics[width=0.92\textwidth]{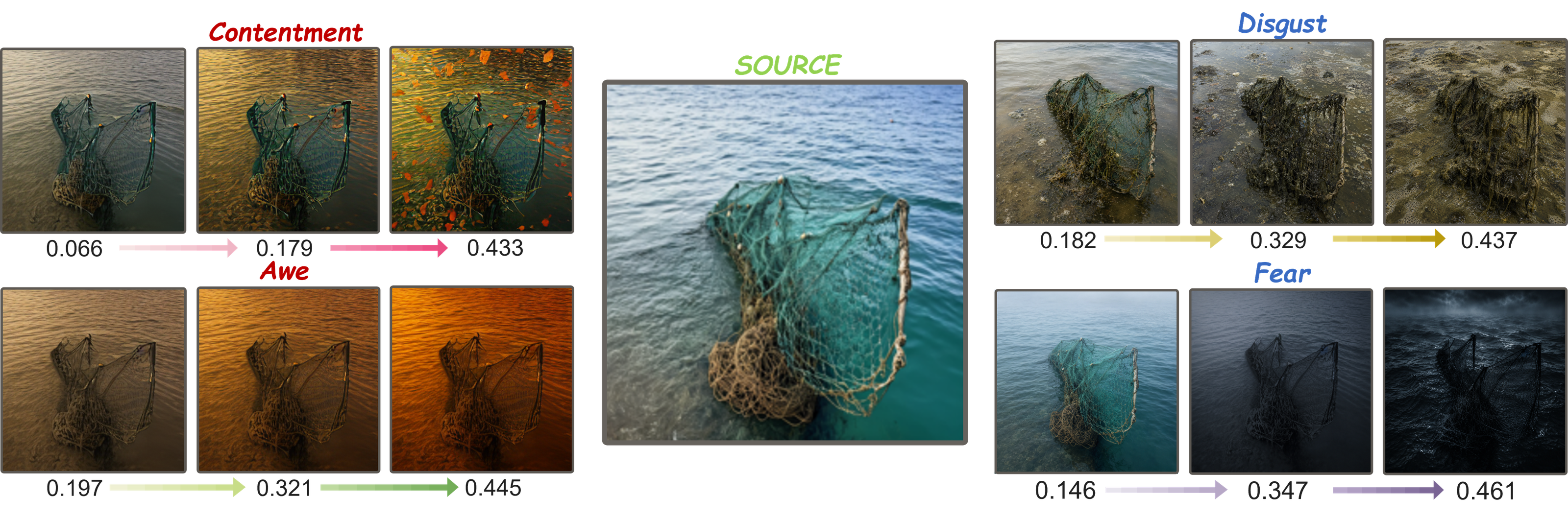}
    \caption{\textbf{Multi-directional and graded affective control.} Starting from the same source image, \methodname{} follows four target emotion directions---\emph{Contentment}, \emph{Awe}, \emph{Disgust}, and \emph{Fear}. Within each direction, varying the target-emotion strength specified by the input distribution generates different images with corresponding affective intensities. Numbers below the outputs denote the corresponding emotion's proportion in the target distribution.}
    \label{fig:graded-control}
    \vspace{-5mm}
\end{figure*}

\subsection{Comparison with Baselines}

\paragraph{Quantitative comparison.}
Table~\ref{tab:baseline-results} shows that \methodname{} ranks first on all six metrics. Its Top-1 of 0.899 and JSD of 0.056 demonstrate accurate control of the dominant emotion and the complete target distribution, respectively, directly supporting our full-distribution formulation. The positive NAP (0.199) and high $\Delta$-Cos (0.744) further show that the outputs follow the requested signed source-to-target transition rather than merely matching a target category. Meanwhile, the highest CLIP-I (0.778) and lowest LPIPS (0.382) indicate that these affective gains are achieved while preserving source content. Together, the results support combining distribution-valued control with source-aware, scene-dependent visual realization.

\begin{table}[htbp]
\centering
\caption{Quantitative comparison on the 4,977-pair test set. Bold and underline denote the best and second-best results, respectively.}
\label{tab:baseline-results}
\scriptsize
\setlength{\tabcolsep}{2.0pt}
\resizebox{\columnwidth}{!}{%
\begin{tabular}{lcccccc}
\toprule
Method & Top-1 $\uparrow$ & JSD $\downarrow$ & NAP $\uparrow$ & $\Delta$-Cos $\uparrow$ & CLIP-I $\uparrow$ & LPIPS $\downarrow$ \\
\midrule

LDAST & 0.267 & 0.324 & $-5.568$ & 0.266 & 0.670 & 0.627 \\
SDEdit & 0.381 & 0.248 & $-3.417$ & 0.245 & 0.631 & 0.603 \\
IP2P & 0.516 & \underline{0.117} & $\underline{-0.095}$ & 0.234 & 0.656 & \underline{0.480} \\
AIF & 0.494 & 0.152 & $-1.163$ & 0.237 & 0.618 & 0.668 \\
EmoEdit & \underline{0.517} & 0.126 & $-0.511$ & \underline{0.316} & \underline{0.687} & 0.513 \\
EmoEditor & 0.453 & 0.201 & $-2.003$ & 0.260 & 0.597 & 0.570 \\
\midrule
\methodname{} (Ours) & \textbf{0.899} & \textbf{0.056} & \textbf{0.199} & \textbf{0.744} & \textbf{0.778} & \textbf{0.382} \\
\bottomrule
\end{tabular}
}
\vspace{-3mm}
\end{table}

\paragraph{Qualitative comparison.}
Figure~\ref{fig:qualitative-comparison} shows that \methodname{} obtains the lowest JSD in all four cases spanning \emph{Amusement}, \emph{Fear}, \emph{Awe}, and \emph{Disgust}. The edits adapt semantic and appearance changes to each source scene rather than applying a fixed text-linked operation.

\subsection{Multi-Directional and Graded Affective Control}
Figure~\ref{fig:graded-control} examines whether a single source image can be steered toward different affective directions and intensities. From the same source, \methodname{} produces distinct trajectories toward \emph{Contentment}, \emph{Awe}, \emph{Disgust}, and \emph{Fear}. Within each trajectory, increasing the target-emotion strength specified by the input distribution generates different images with progressively stronger affective expression while preserving the underlying scene. This shows that the complete target distribution provides both an affective direction and fine-grained control over its intensity.

\subsection{Ablation Study}

\begin{table}[htbp]
\centering
\caption{Ablation study on the 4,977-pair test set. The complete distribution difference yields the strongest affective alignment. Bold and underline denote the best and second-best results, respectively.}
\label{tab:ablation-results}
\scriptsize
\setlength{\tabcolsep}{1.6pt}
\resizebox{\columnwidth}{!}{%
\begin{tabular}{lcccccc}
\toprule
Variant & Top-1 $\uparrow$ & JSD $\downarrow$ & NAP $\uparrow$ & $\Delta$-Cos $\uparrow$ & CLIP-I $\uparrow$ & LPIPS $\downarrow$ \\
\midrule
One-hot Target & 0.642 & 0.122 & $-0.121$ & 0.293 & 0.653 & 0.519 \\
Target Only & 0.761 & \underline{0.069} & 0.023 & \underline{0.694} & \textbf{0.794} & \underline{0.352} \\
w/o $\mathcal{L}_{\mathrm{emo}}$ & \underline{0.826} & 0.073 & \underline{0.129} & 0.651 & 0.721 & \textbf{0.328} \\
+ Text Alignment & 0.768 & 0.094 & 0.086 & 0.597 & 0.743 & 0.435 \\
\midrule
\methodname{} (Ours) & \textbf{0.899} & \textbf{0.056} & \textbf{0.199} & \textbf{0.744} & \underline{0.778} & 0.382 \\
\bottomrule
\end{tabular}
}
\vspace{-3mm}
\end{table}

Table~\ref{tab:ablation-results} supports the central design choices of \methodname{}. Replacing the target distribution with a one-hot category degrades Top-1 (0.899 to 0.642), JSD (0.056 to 0.122), and $\Delta$-Cos (0.744 to 0.293), showing that category-level control loses information needed for distributional transitions. Target-only conditioning also trails the signed source-to-target difference in NAP (0.023 vs.\ 0.199) and $\Delta$-Cos (0.694 vs.\ 0.744), confirming the importance of the source affective state. Removing $\mathcal{L}_{\mathrm{emo}}$ or adding text alignment lowers $\Delta$-Cos to 0.651 and 0.597, respectively, supporting direct distribution-transition supervision. The lower LPIPS of Target Only and w/o $\mathcal{L}_{\mathrm{emo}}$ coincides with weaker affective scores, indicating more conservative rather than better-controlled edits.

\section{Conclusion}
We presented \methodname{}, a source-aware editor driven by signed transitions between complete eight-dimensional emotion distributions. Trained on the 248,841-pair \datasetname{} without editing-text supervision, it learns scene-dependent visual realizations of affective change. Comparisons with six baselines, ablations, and graded-control results show improved affective alignment and content preservation, validating distribution-valued control for fine-grained emotional image editing.

\bibliography{aaai2027}\clearpage\appendix\twocolumn[\begin{center}{\LARGE\bfseries Supplementary Material\par}\vspace{0.75em}\end{center}]

\section{Evaluation Metrics}
\label{sec:supp-metrics}

Let $\mathbf{q}_s,\mathbf{q}_t\in[0,1]^8$ denote the reference emotion distributions stored in the evaluation metadata for the source and target images, and let $\widehat{\mathbf{p}}_s,\widehat{\mathbf{p}}_e\in[0,1]^8$ denote the distributions predicted by the independent evaluator for the source and edited images. The evaluator rejects non-finite, negative, or zero-mass vectors and normalizes every accepted vector to unit mass. We use $\epsilon=10^{-12}$ for numerical stability. Each metric is computed for every successfully evaluated test record, and the reported result is the arithmetic mean over these records. The implementation supports either CLIP-8 or EDP as the emotion backend; the selected backend and model identity are stored in the evaluation configuration.

\subsection{Affective Alignment}

\paragraph{Target Top-1 accuracy.}
Target Top-1 measures whether the dominant emotion predicted for the edited image agrees with the dominant component of the target reference distribution:
\begin{equation}
\mathrm{Top\mbox{-}1}
=
\mathbf{1}\!\left[
\underset{k}{\arg\max}\;q_{t,k}
=
\underset{k}{\arg\max}\;\widehat{p}_{e,k}
\right].
\end{equation}
Higher is better.

\paragraph{Jensen--Shannon divergence (JSD).}
The code uses base-2 JSD to measure full-distribution discrepancy:
\begin{align}
\mathbf{m} &= \tfrac{1}{2}(\mathbf{q}_t+\widehat{\mathbf{p}}_e),\\
D_{\mathrm{JS}}^{(2)}(\mathbf{q}_t,\widehat{\mathbf{p}}_e)
&=
\tfrac{1}{2}D_{\mathrm{KL}}^{(2)}(\mathbf{q}_t\|\mathbf{m})
+
\tfrac{1}{2}D_{\mathrm{KL}}^{(2)}(\widehat{\mathbf{p}}_e\|\mathbf{m}),\\
D_{\mathrm{KL}}^{(2)}(\mathbf{p}\|\mathbf{q})
&=
\sum_{k:p_k>0}p_k\log_2\frac{p_k}{q_k}.
\end{align}
Zero-probability terms are omitted. The final JSD is clipped to $[0,1]$ only to suppress floating-point roundoff; lower is better.

\paragraph{Normalized Affective Progress (NAP).}
NAP measures progress toward the target relative to the independently predicted source state. We define
\begin{align}
d_s &= D_{\mathrm{JS}}^{(2)}(\widehat{\mathbf{p}}_s,\mathbf{q}_t), &
d_e &= D_{\mathrm{JS}}^{(2)}(\widehat{\mathbf{p}}_e,\mathbf{q}_t),\\
&\mathrm{NAP}
=
\frac{d_s-d_e}{d_s+\epsilon}.
\end{align}
NAP approaches $1$ when the edited distribution reaches the target, equals $0$ when editing makes no progress, and becomes negative when the output moves away from the target. The metric is not clipped, so a small $d_s$ can amplify negative values.

\paragraph{Transition-direction cosine ($\Delta$-Cos).}
This metric compares the requested transition between the stored reference distributions with the transition predicted by the independent evaluator:
\begin{align}
\boldsymbol{\delta}_{r}
&=
\mathbf{q}_t-\mathbf{q}_s, 
\\\boldsymbol{\delta}_{p}
&=
\widehat{\mathbf{p}}_e-\widehat{\mathbf{p}}_s,\\
c_{\Delta}
&=
\frac{
\boldsymbol{\delta}_{r}^{\top}\boldsymbol{\delta}_{p}
}{
\|\boldsymbol{\delta}_{r}\|_2\|\boldsymbol{\delta}_{p}\|_2+\epsilon
},\\
\Delta\text{-Cos}
&=
\operatorname{clip}(c_{\Delta},-1,1).
\end{align}
Higher is better. If either transition has zero norm, the numerator is zero and the implementation records $0$ rather than excluding the instance.

\subsection{Content Preservation}

\paragraph{CLIP image similarity (CLIP-I).}
Let $g(\cdot)$ be the CLIP image encoder. The implementation $\ell_2$-normalizes its image features and computes
\begin{equation}
\mathrm{CLIP\mbox{-}I}
=
\operatorname{clip}\!\left(
\frac{g(I_s)^{\top}g(I_e)}
{\|g(I_s)\|_2\,\|g(I_e)\|_2},
-1,1
\right),
\end{equation}
where higher values indicate stronger semantic-content preservation. The default encoder is \url{openai/clip-vit-large-patch14}.

\paragraph{Learned perceptual image patch similarity (LPIPS).}
The evaluator uses the standard pretrained LPIPS model with an AlexNet backbone. Each RGB image is resized to $256\times256$ with Lanczos resampling and scaled from $[0,255]$ to $[-1,1]$. For normalized feature maps $\widehat{f}_l$ and learned channel weights $\boldsymbol{\omega}_l$, the corresponding distance is
\begin{align}
\mathbf{r}_{l,h,w}
&=
\widehat{f}_l(I_s)_{h,w}-\widehat{f}_l(I_e)_{h,w},\\
\mathrm{LPIPS}
&=
\sum_l\frac{1}{H_lW_l}\sum_{h,w}
\left\|\boldsymbol{\omega}_l\odot\mathbf{r}_{l,h,w}\right\|_2^2,
\end{align}
where lower is better. CLIP-I and LPIPS use identical model configurations and metric-specific preprocessing for every compared method.

\section{Implementation Details}
\label{app:Implementation_Details}

\subsection{EDP}
We implement EDP as an eight-output ResNet-18 initialized from the emotion classifier used by EmoEditor~\cite{lin2025make,he2016deep}. We fine-tune its final residual stage and prediction head on the 16,957 training images from Flickr-LDL and Twitter-LDL and use their 4,238 held-out images for model selection~\cite{yang2017learning,yang2017joint}. The LDL target vectors are remapped to the common eight-emotion order before training. Images are resized to $224\times224$. We minimize batch-mean KL divergence to the vote-derived distribution with AdamW for 10 epochs using a learning rate and weight decay of $10^{-4}$ and cosine annealing over all epochs. Training uses automatic mixed precision, eight data-loading workers, natural sampling without distillation, and seed 75863. 

\subsection{AffectDelta}
We train \methodname{} on 243,864 source-group-disjoint image pairs at $224\times224$ resolution. The four-layer transition encoder maps each eight-dimensional signed emotion difference to 77 tokens of width 768, matching the cross-attention interface of the pretrained InstructPix2Pix backbone~\cite{brooks2023instructpix2pix}. We jointly optimize the transition encoder and U-Net while freezing the VAE and EDP. AdamW uses a learning rate of $10^{-5}$, weight decay $0$, $\beta=(0.9,0.999)$, and $\epsilon=10^{-8}$ for 15 epochs, with batch size 32 per GPU, one gradient-accumulation step, no learning-rate scheduler, and gradient clipping at norm 1.0. Training uses distributed data parallelism, FP16 automatic mixed precision, channels-last tensors, 16 data-loading workers with prefetch factor 2, and seed 75863; thus, the effective global batch size is $32$ times the number of GPUs. The objective is $\mathcal{L}_{\mathrm{diff}}+0.1\mathcal{L}_{\mathrm{emo}}$. The diffusion term is the standard noise-prediction MSE with timesteps sampled uniformly over the pretrained scheduler. At every optimizer step, the emotion term samples four examples with inverse target-class-frequency weights, draws low-noise timesteps uniformly from $[0,250]$, reconstructs a one-step clean latent clipped to $[-10,10]$, and minimizes MSE between the frozen EDP prediction and the target distribution. 

\section{EDP Fine-Tuning Analysis}
\label{app:edp-analysis}

We evaluate the original single-label initialization and the distribution-fine-tuned EDP on the same 4,238 held-out Flickr-LDL and Twitter-LDL images. Top-1 measures agreement between $\arg\max_k q_k$ and $\arg\max_k \widehat{p}_k$; KL and MAE measure full-distribution error, with $\mathrm{MAE}=\frac{1}{8}\|\mathbf{q}-\widehat{\mathbf{p}}\|_1$. Mean max probability is $\frac{1}{N}\sum_i\max_k\widehat{p}_{i,k}$ and measures prediction sharpness rather than accuracy.

\begin{table*}[t]
\centering
\small
\caption{Held-out evaluation of EDP fine-tuning on 4,238 distribution-labeled images. Fine-tuning improves both dominant-emotion agreement and full-distribution fidelity; the lower mean maximum probability indicates reduced over-concentration rather than degraded accuracy.}
\label{tab:edp-finetuning}
\setlength{\tabcolsep}{9pt}
\begin{tabular}{lccccc}
\toprule
Predictor & $N$ & Top-1 $\uparrow$ & KL $\downarrow$ & MAE $\downarrow$ & Mean max prob. \\
\midrule
Before distribution fine-tuning & 4,238 & 0.2298 & 3.6682 & 0.1745 & 0.7626 \\
After distribution fine-tuning & 4,238 & \textbf{0.7964} & \textbf{0.3456} & \textbf{0.0522} & 0.4532 \\
\bottomrule
\end{tabular}
\end{table*}

\paragraph{Metric analysis.}
Fine-tuning raises Top-1 agreement from 0.2298 to 0.7964 while reducing KL from 3.6682 to 0.3456 and MAE from 0.1745 to 0.0522. The initialization's high mean maximum probability (0.7626) follows from its eight-way single-label objective, which rewards concentrating probability mass on one category even when annotator votes support several emotions. Distribution-level supervision lowers this value to 0.4532 while improving both dominant-class agreement and full-distribution fidelity. The joint improvement shows that the change is not indiscriminate smoothing; it instead indicates reduced single-label overconfidence together with recovery of plausible non-dominant probability mass.

\begin{figure*}[t]
\centering
\includegraphics[width=0.98\textwidth]{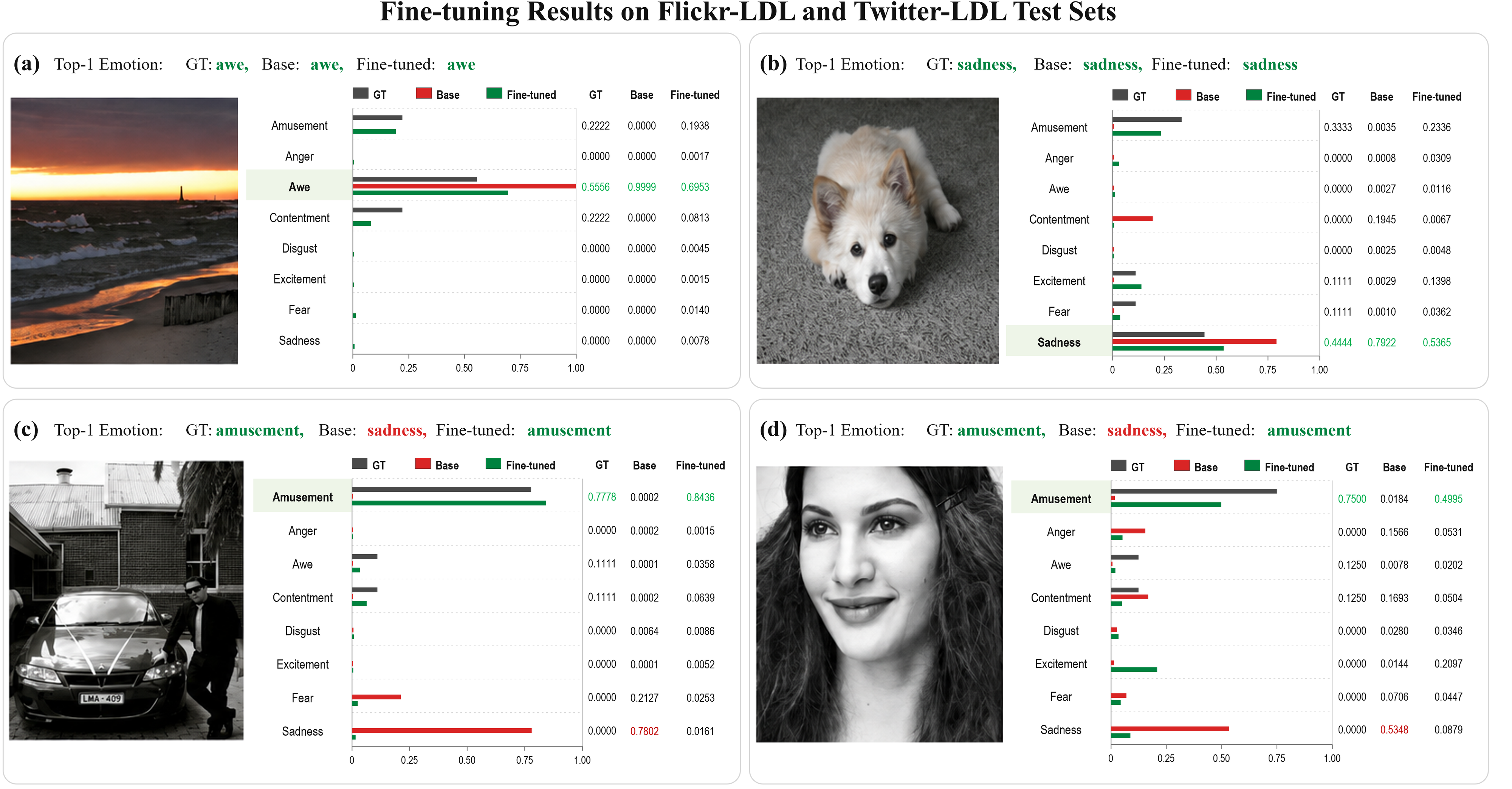}
\caption{Example-level verification of EDP fine-tuning on Flickr-LDL and Twitter-LDL. Gray bars denote vote-derived targets, red bars the single-label initialization, and green bars the fine-tuned EDP. In (a) and (b), both predictors match the target Top-1 emotion, but the initialization collapses most mass onto that class; fine-tuning better recovers secondary emotions. In (c) and (d), fine-tuning corrects the initialization's spurious sadness predictions on grayscale amusement examples.}
\label{fig:edp-finetuning-examples}
\end{figure*}

\paragraph{Visual analysis.}
Figure~\ref{fig:edp-finetuning-examples} complements the aggregate metrics with example-level distributions. In panels (a) and (b), both predictors match the vote-derived Top-1 emotion, but the single-label initialization is sharply concentrated. It assigns 0.9999 to awe in (a), although the target assigns 0.5556 to awe and 0.2222 to both amusement and contentment; in (b), it assigns 0.7922 to sadness, whereas the target assigns 0.4444 to sadness together with 0.3333 amusement and 0.1111 each for excitement and fear. Fine-tuning preserves the correct Top-1 decisions while recovering much of this secondary mass. Panels (c) and (d) expose a complementary failure mode: the initialization predicts sadness (0.7802 and 0.5348) for two grayscale images whose vote-derived Top-1 emotion is amusement, whereas the fine-tuned EDP restores amusement as Top-1 (0.8436 and 0.4995) and suppresses the spurious sadness mass. These examples are consistent with sensitivity to low-level appearance rather than affective semantics. In affective editing, global darkening, red hue shifts, contrast amplification, or heavy stylization can exploit such sensitivity and raise a target score without a commensurate semantic change; cross-test analyses likewise report high automatic scores for color- or style-dominated edits, including broad overlays that obscure original content~\cite{lin2025make}. Taken together, the quantitative gains and corrected examples show that distribution-level fine-tuning solves this shortcut-driven failure of the original classifier, reducing its reliance on low-level appearance cues and yielding emotion distributions that better reflect affective semantics.

\section{Dataset Construction and Splits}
\paragraph{Source-grouped partition.}
Following the transition-wise organization of EmoEdit and EmoEditor~\cite{yang2025emoedit,lin2025make}, we perform source grouping before determining the final retained dataset. We place candidate pairs in the same group whenever they share an original source identifier, one pair's target reappears as another pair's source, their images have the same content hash, they are perceptual near-duplicates produced by resizing, cropping, or compression, or they are alternative instructions or generated versions derived from the same source image. After taking the transitive closure of these relations and completing task-specific filtering, 248,841 pairs remain. We then assign each complete group wholly to either training or test, preventing source identities, edit chains, and near-duplicate images from crossing the split.

\paragraph{Balanced transition core.}
The test set contains a strictly balanced 4,960-pair core organized over all 64 directed Top-1 transition units, as summarized in Table~\ref{tab:test-split}. The 56 off-diagonal units contain 60 pairs each, equally divided between GPT-Image-Edit-1.5M and ImgEdit, while each of the eight same-Top-1 units contains 200 pairs, with 100 from each source dataset. The core therefore contains 620 pairs for every source Top-1 emotion and 620 for every target Top-1 emotion; its 1,600 same-Top-1 pairs account for 32.3\% of the core. Because source groups are indivisible, group-complete assignment contributes 17 additional audit pairs beyond these exact quotas, yielding the final 4,977-pair test set and leaving 243,864 pairs for training. Exact balance statements refer to the 4,960-pair core, whereas all reported evaluation metrics use all 4,977 test pairs.

\begin{center}
\small
\captionof{table}{Strictly balanced core of the test split.}
\label{tab:test-split}
\begin{tabular}{@{}cccc@{}}
\toprule
Transition type & Units & Per unit & Pairs \\
\midrule
Cross-valence, different Top-1 & 32 & 60 & 1,920 \\
Same-valence, different Top-1 & 24 & 60 & 1,440 \\
Same Top-1, distribution shift & 8 & 200 & 1,600 \\
\midrule
Total & 64 & -- & 4,960 \\
\bottomrule
\end{tabular}
\end{center}

\paragraph{Within-unit stratification.}
Same-Top-1 candidates must satisfy the fixed base-2 criterion $D_{\mathrm{JS}}^{(2)}(\mathbf{p}_s,\mathbf{p}_t)\geq\tau_{\mathrm{JSD}}$. Within every transition unit, we then sample by empirical low, medium, and high strata of base-2 JSD or $\|\Delta\mathbf{p}\|_1$, target-distribution entropy, and scene type, including people, animals, natural landscapes, indoor scenes, cities, and objects. Same-Top-1 units additionally balance three transition patterns: strengthening the dominant emotion, weakening it while retaining the same Top-1 category, and approximately preserving it while redistributing secondary emotions. All random choices use the fixed split seed $s_{\mathrm{split}}$.

\end{document}